\documentclass{article}
\usepackage{spconf,amsmath,epsfig}

\let\OLDthebibliography\thebibliography
\renewcommand\thebibliography[1]{
  \OLDthebibliography{#1}
  \setlength{\parskip}{0pt}
  \setlength{\itemsep}{0pt plus 0.3ex}
}

\begin{document}\sloppy

\def\x{{\mathbf x}}
\def\L{{\cal L}}

\title{Composite Localization for Human Pose Estimation}
%
\name{ZiFan Chen, Xin Qin, Chao Yang, Li Zhang}
\address{}

\maketitle

\begin{abstract}
The existing human pose estimation methods are confronted with inaccurate long-distance regression or high computational cost due to the complex learning objectives. This work proposes a novel deep learning framework for human pose estimation called composite localization to divide the complex learning objective into two simpler ones: a sparse heatmap to find the keypoint's approximate location and two short-distance offsetmaps to obtain its final precise coordinates. To realize the framework, we construct two types of composite localization networks: CLNet-ResNet and CLNet-Hourglass. We evaluate the networks on three benchmark datasets, including the Leeds Sports Pose dataset, the MPII Human Pose dataset, and the COCO keypoints detection dataset. The experimental results show that our CLNet-ResNet50 outperforms SimpleBaseline by 1.14\% with about 1/2 GFLOPs. Our CLNet-Hourglass outperforms the original stacked-hourglass by 4.45\% on COCO.
\end{abstract}
%

\section{Introduction}
\label{sec:intro}
Human pose estimation, predicting a person's body part or joint positions from an image or a video, is fundamental in computer vision with plenty of applications in human-computer interaction, action recognition, and other practical tasks. Recently, deep neural networks have surpassed the previous methods based on hand-crafted features by significantly improving the prediction accuracy in human pose estimation~\cite{toshev2014deeppose,Papandreou_2017_CVPR,xiao2018simple, Sun_2019_CVPR}. 

\begin{figure}[t]
	\begin{center}
		\includegraphics[width=0.90\linewidth]{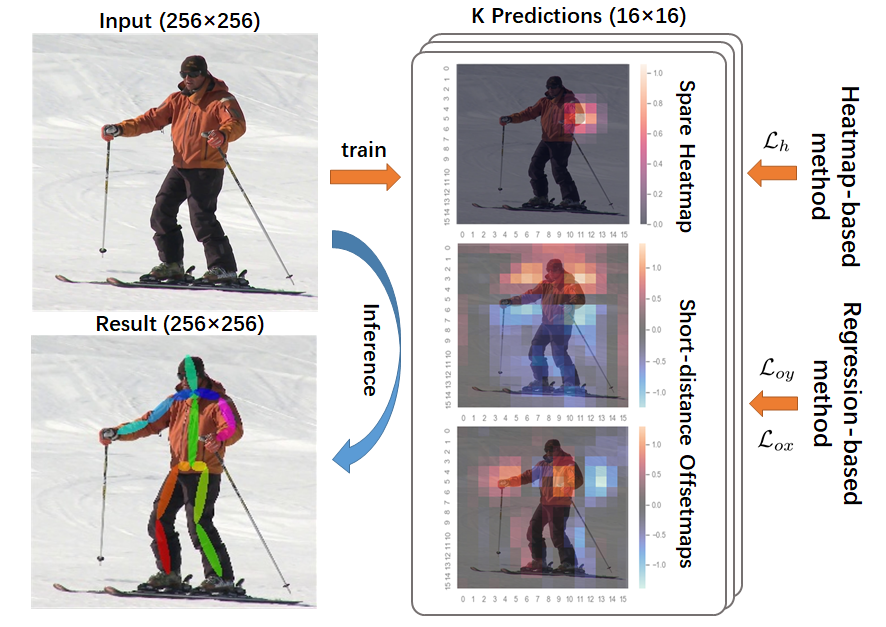}
		
	\end{center}
	\caption{Illustration of composite localization framework.}
	\label{fig:framework}
\end{figure}

The human pose estimation based on deep learning can be divided into regression-based and heatmap-based methods. The regression-based method can predict the coordinates of keypoints in an end-to-end fashion but may sacrifice prediction accuracy due to the long-range information in the whole image~\cite{toshev2014deeppose,carreira2016human,sun2017compositional}. The heatmap-based method predicts the probability of different keypoints on specific pixels and forms a heatmap to present the probabilities~\cite{tompson2014joint,Tompson_2015_CVPR,newell2016stacked,Yang_2017_ICCV,Tang_2018_ECCV,Tang_2019_CVPR}, which usually produce higher prediction accuracy but require more computational resources for predicting the dense heatmap. Both types of methods have merits and limitations, such as inaccurate regression or high computational cost and complexity due to the complex learning objectives like long-distance regression or dense heatmap prediction. There are also a handful of studies that have attempted to combine the two types of methods, but they often fail to achieve satisfactory accuracy because they also ignore the complexity of learning objectives~\cite{Sun_2018_ECCV}.

\begin{figure*}
	\begin{center}
		\includegraphics[width=0.85\linewidth]{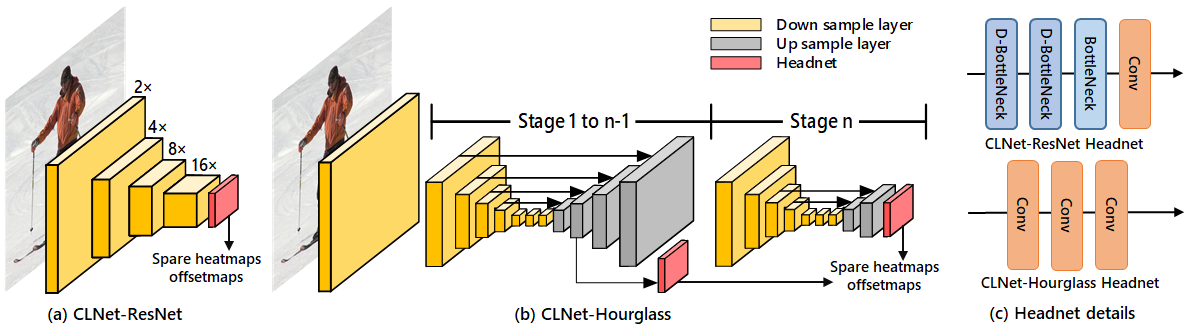}
	\end{center}
	\caption{Illustration of CLNet-ResNet and CLNet-Hourglass. "D-BottleNeck" stands for the dilated bottleneck in ~\cite{Li_2018_ECCV}.}
	\label{fig:model}
\end{figure*}

In this work, we find that the coordinates of keypoints can be divided into two simpler expressions: an approximate location and the corresponding short-distance regression. Accordingly, using a low-resolution heatmap (sparse heatmap) represents approximate locations to narrow the regression range. At the same time, regression-based method can be carried out short-distance regression from the approximate position. Such two simple objectives are suitable for neural networks to learn rather than a complex one~\cite{ke2020green}. Based on this, we propose a composite localization framework to predict sparse heatmaps and short-distance offsetmaps simultaneously. The main contributions can be summarized as follows: 1) We propose a composite localization framework (CL) for human pose estimation based on the two simple objectives and design the appropriate loss to improve performance. 2) We construct two types of composite localization networks, CLNet-ResNet and CLNet-Hourglass, based on our modified ResNet and Hourglass to show that our framework can be simply added to an existing model. 3) We evaluate CLNets on three benchmark datasets and show that CLNets achieve start-of-the-art performance and prove the rationality of framework design by sufficient ablation experiments.

\section{Related Work}
\label{sec:related}

Early studies of human pose estimation have limited practical applications, primarily because they rely heavily on hand-crafted features~\cite{yang2012articulated}. Most recent work is based on deep learning and can be roughly divided into regression-based and heatmap-based methods. There are also some works trying to combine these two ideas for better performance.

\noindent\textbf{Regression-based Method} 
The first human pose estimation based on deep learning, DeepPose~\cite{toshev2014deeppose}, proposes a cascaded deep neural network with extracting information evenly from the whole image to regress keypoints. Carreira et al.~\cite{carreira2016human} use a self-correcting model to expand the expression ability of hierarchical feature extractors. Sun et al.~\cite{sun2017compositional} use bones to reparameterize the pose representation and joint connection structure to encode the long-range interactions in the specific posture. Although these methods improve the regression accuracy, they are still learning the complex goal of long-distance regression.

\noindent\textbf{Heatmap-based Method} 
Shortly after DeepPose was published, Tompson et al.~\cite{Tompson_2015_CVPR} use heatmaps to represent the probabilities of keypoints in different locations. The stacked Hourglass architecture proposed by Newell et al.~\cite{newell2016stacked} uses repeated encode-decode structures with multiple supervision on intermediate heatmaps to improve the accuracy of the final prediction results. Many models~\cite{Yang_2017_ICCV,Ke_2018_ECCV,Tang_2018_ECCV,Tang_2019_CVPR} continuously improve the performance of the classic stacked Hourglass network. Xiao et al.~\cite{xiao2018simple} propose simple but effective baseline methods, named SimpleBaseline. Chen et al.~\cite{chen2018cascaded} use a two-stage strategy to further optimize the model for difficult samples. Sun et al.~\cite{Sun_2019_CVPR} maintain the high resolution of the model by training multi-resolution subnetworks. Although these models have good performance, learning complex dense heatmaps require complex network architecture and computational cost.

\noindent\textbf{Composite Method}
A handful of works are also introduced to combine the regression-based and heatmap-based methods to overcome these methods' shortcomings. Sun et al.~\cite{Sun_2018_ECCV} estimate the positions of keypoints as the integrals of all positions in the heatmaps to preserve the end-to-end differentiability. Papandreou et al.~\cite{Papandreou_2017_CVPR} solve a binary classification problem for each position, and all the positive locations need to predict offsets towards the keypoints.

\section{Proposed Method}
\label{sec:cl_method}
\subsection{Composite Localization}
As shown in Figure~\ref{fig:framework}, the composite localization framework uses a sparse heatmap to find an approximate position and two corresponding offsetmaps to carry out short-distance regression.

Suppose the original image size is $W \times H$, the number of keypoints is $K$, and the sizes of sparse heatmaps and short-distance offsetmaps are $W' \times H'$. There are the following relationships: $W'=\left \lfloor W/S \right \rfloor$ and $H'=\left \lfloor H/S \right \rfloor$, where $S$ is the downsampling stride. Thus, each location in sparse heatmaps or short-distance offsetmaps corresponds to a patch of the original image with $S \times S$ size.   

\noindent\textbf{Sparse Heatmap} For $K$ keypoints, there are $K$ sparse heatmaps, $\{\mathbf{H_1}, \mathbf{H_2}, ..., \mathbf{H_K}\}$. Suppose the ground truth location of the $k$th keypoint in the original image is defined as $\mathbf{g}^k=(g_x^k,g_y^k), g_x^k \in \{1...W\}, g_y^k \in \{1...H\}$. The value at $\mathbf{p'}=(p'_x,p'_y), p'_x \in \{1...W'\}, p'_y \in \{1...H'\}$ in $\mathbf{H_k}$ is defined as,
\begin{align}
\mathbf{H_k}(\mathbf{p'})=exp\begin{pmatrix}\frac{-\left \| t(\mathbf{p'})-\mathbf{g}^k \right \|_2}{2\sigma ^2}\end{pmatrix},
\end{align}
where $t(\mathbf{p'})$ translates the location $\mathbf{p'}=(p'_x,p'_y)$ in $\mathbf{H_k}$ to the center coordinates of the corresponding patch in the original image, which can be expressed as,
\begin{align}
t(\mathbf{p'})=\begin{pmatrix}(p'_x-C) \times S,(p'_y-C) \times S \end{pmatrix},
\end{align} 
where $C$ is a deviation constant, equals to 0.5.

\begin{table*}
	\begin{center}
		\resizebox{0.98\textwidth}{!}{
			\begin{tabular}{l|c|c|c|c|cccccc|ccccccc}
				\hline
				 & & & & & \multicolumn{6}{|c|}{val2017} & \multicolumn{6}{|c}{test-dev2017}\\
				Method &Backbone &Pretrain &Input Size & GFLOPs & AP &$AP^{50}$ &$AP^{75}$ & $AP^{M}$ &$AP^{L}$ & AR &AP &$AP^{50}$ &$AP^{75}$ & $AP^{M}$ &$AP^{L}$ & AR \\
				\hline
				\hline
				Hourglass ~\cite{newell2016stacked}
				& 8 stacked hourglass & N & 256$\times$192 & 14.3 & 71.9 &91.0 & 80.0 & 69.3 & 77.1 & 77.5 & - & - & - & - & - & - \\
				CPN~\cite{chen2018cascaded}
				& ResNet50 & Y & 256$\times$192 & 6.2 & 69.2 & 88.0 & 76.2 & 65.8 & 75.6 & - & - & - & - & - & - & - \\
				SimpleBaseline~\cite{xiao2018simple}
				& ResNet50 & Y & 256$\times$192 & 8.9 & 70.4 & 88.6 & 78.3 & 67.1 & 77.2 & 76.3 & - & - & - & - & - & - \\
				HRNet~\cite{Sun_2019_CVPR}
				& HRNet-W48 & Y & 256$\times$192 & 14.6 & \textbf{75.1} & 90.6 & 82.2 & 71.5 & 81.8 & 80.4 & - & - & - & - & - & - \\
				Integral Pose Regression~\cite{Sun_2018_ECCV} & ResNet101 & Y & 256$\times$256 & 11.0 & - & - & - & - & - & - & 67.8 & 88.2 & 74.8 & 63.9 & 74.0 & - \\
				Cai et al.~\cite{carreira2016human} & ResNet50 & N & 256$\times$192 & 6.4 & 74.7 & 91.4 & 81.5 & 71.0 & 80.2 & 80.0 & 72.5 & 93.0 & 81.3 & 69.9 & 76.5 & 78.8 \\
				\hline
				CLNet-ResNet 
				& ResNet50 & Y &256$\times$192 & \textbf{4.2} & 71.2 & 	88.8&78.5&67.4&77.8&78.2& - & - & - & - & - & - \\
				CLNet-Hourglass 
				& 8 stacked hourglass & N &  256$\times$192 & 26.5 &\textbf{75.1}&89.4&81.8 &71.7&81.6&82.0& - & - & - & - & - & - \\
				
				\hline
				\hline
				G-RMI~\cite{Papandreou_2017_CVPR} & ResNet101 & Y & 353$\times$257 & 57.0 & - & - & - & - & - & - & 64.9 & 85.5 & 71.3 & 62.3 & 70.0 & 69.7\\
				CPN~\cite{chen2018cascaded}
				& ResNet-Inception & Y & 384$\times$288 & - & 72.2&89.2&78.6&68.1&79.3& - & 72.1 & 91.4 & 80.0 & 68.7 & 77.2 & 78.5 \\
				SimpleBaseline~\cite{xiao2018simple}
				& ResNet152 & Y & 384$\times$288 & 35.6 &74.3&89.6&81.1&70.5&81.6&79.7&73.7&91.9&82.8&71.3&80.0&79.0\\
				
				HRNet~\cite{Sun_2019_CVPR}
				& HRNet-W32 & Y & 384$\times$288 & 16.0  &75.8&90.6&82.5&72.0&82.7&80.9& 74.9 & 92.5 & 82.8 & 71.3 & 80.9 & 80.1 \\
				HRNet~\cite{Sun_2019_CVPR}
				& HRNet-W48 & Y & 384$\times$288 & 32.9 & 76.3	&90.8&82.9&72.3&83.4&81.2&75.5&92.5&83.3&71.9&81.5&80.5 \\
				\hline
				CLNet-ResNet 
				& ResNet50 & Y & 384$\times$288 & \textbf{9.5} & 73.4&89.1&79.9&69.4&80.0&79.7& - & - & - & - & - & - \\
				CLNet-ResNet 
				& ResNet101 & Y & 384$\times$288 & 17.7 & 74.2&89.3&80.5&70.2&81.0&80.5& - & - & - & - & - & - \\
				CLNet-ResNet 
				& ResNet152 & Y & 384$\times$288 & 25.9 & 
				74.9 & 89.7 & 81.4 & 70.9 & 81.9 & 81.1&74.1&91.5&81.5&70.4&80.1&80.3\\
				CLNet-Hourglass 
				& 8 stacked hourglass&N & 384$\times$256& 52.9 & \textbf{76.5}&89.8&82.8&72.9&82.9&82.6&\textbf{75.8}&91.7&83.2&72.4&81.4&82.0  \\
				\hline
			\end{tabular}
		}
	\end{center}
	\caption{Comparisons of results on COCO val2017 and test-dev2017 set.}
	\label{table:COCO val test}
\end{table*}

\noindent\textbf{Short-distance Offsetmaps} For $K$ keypoints, there are $2K$ offsetmaps, $\{\mathbf{O_1}, \mathbf{O_2}, ..., \mathbf{O_K}, \mathbf{O_{K+1}}, ..., \mathbf{O_{2K}}\}$, where the first $K$ and the last $K$ offsetmaps predict y-offsets and x-offsets, respectively.
Similarly, for the ground truth location $\mathbf{g}^k$, the value at $\mathbf{p'}=(p'_x,p'_y)$ in $\mathbf{O_k}$ and $\mathbf{O_{K+k}}$ can be defined as,
\begin{align}
\left\{\begin{matrix}
\mathbf{O_k}(\mathbf{p'})&=\left (g_y^k-(p'_y-C)\times S \right ) / S, \\
\mathbf{O_{K+k}}(\mathbf{p'})&=\left (g_x^k-(p'_x-C)\times S \right ) / S, 
\end{matrix}\right.
\end{align}
where $C$ and $S$ have the same meaning as above.

\subsection{Network Design} We construct two networks to test the effectiveness and generalizability of our CL framework: CLNet-ResNet and CLNet-Hourglass.

CLNet-ResNet uses ResNet's first four feature extraction stages~\cite{he2016deep} as its network backbone, as shown in Figure~\ref{fig:model} (a). Inspired by Detnet~\cite{Li_2018_ECCV}, we sequentially add two dilated bottleneck layers, one bottleneck layer, and one convolutional layer to build the head subnetwork in CLNet-ResNet.

CLNet-Hourglass uses the classic stacked Hourglass~\cite{newell2016stacked} as its network backbone. As shown in Figure~\ref{fig:model} (b), CLNet-Hourglass removes some up-sampling layers in the final stage and generates the predictions with appropriate resolution through the head subnetwork in each stage. The head subnetwork consists of three convolutional layers in series.

\subsection{Loss Function Design} 
For sparse heatmap, let $\mathbf{H}_k$ and $\mathbf{\hat{H}}_k$ be the $k$th ground truth target and the $k$th keypoint prediction, respectively. The loss could be defined as,
\begin{align}
\mathcal L_h=\frac{1}{K}\sum_{k=1}^K f(\mathbf{H}_k,\hat{\mathbf{H}}_k),
\end{align}
where $f(\cdot)$ is the mean square error loss.

Offsetmaps only needs to learn about short-distance regression within a general region provided by sparse heatmaps, and its loss function can be expressed as follows,
\begin{align}
\left\{\begin{matrix}\mathcal L_{oy}=\frac{1}{K}\sum_{k=1}^K\left (\frac{1}{N_{\mathbf{\Omega}}} \sum_{p'\in\mathbf{\Omega}} g(\mathbf{O}_k(g'),\hat{\mathbf{O}}_k(g'))\right ),\\ 
\mathcal L_{ox}=\frac{1}{K}\sum_{k=1}^K\left ( \frac{1}{N_{\mathbf{\Omega}}} \sum_{p'\in\mathbf{\Omega}} g(\mathbf{O}_{K+k}(g'),\hat{\mathbf{O}}_{K+k}(g'))\right ),
\end{matrix}\right.
\end{align}
where $\mathbf{\Omega}$ indicates $\hat{\mathbf{H}_k}(p')\geq \tau$ with using threshold $\tau$ to control the range of regression. $g(\cdot)$ represents the smooth L1 loss. Note that the region may contain the approximate location of the keypoint and the location around it, which allows short-regression from adjacent approximate locations to improve the robustness of the model.

The overall loss function is,
\begin{align}
\mathcal L=\omega_h \mathcal L_h+\mathcal \omega_o (\mathcal L_{oy}+\mathcal L_{ox}),
\end{align}
where $\omega_h$ and $\omega_o$ are the two parts' weights..

\subsection{Inference}
There are three steps to parse sparse heatmaps and short-distance offsetmaps into 2d coordinates vector. For the $k$th keypoint, 1) the locations with high activation value ($\geq \tau$) in $\mathbf{H}_k$ were selected, and their center points were used as the initial locations of regression. 2) The values of $\mathbf{O}_k$ and $\mathbf{O}_{K+k}$ in the corresponding locations are used as y-offset and x-offset to obtain the $k$th keypoint's coordinates. 3) The predicted coordinates are weighted average according to their activation values on $\mathbf{H}_k$ to obtain a final predicted coordinate about the $k$th keypoint.

\section{Experiments}

\begin{table}
	\begin{center}
		\resizebox{0.48\textwidth}{!}{
			\begin{tabular}{lcccccccc}
				\hline
				& Head &Sho. &Elb. &Wri. &Hip &Knee& Ank. &PCKh@0.5 \\
				\hline
				Tompson et al.~\cite{tompson2014joint} 
				& 95.8 & 90.3 & 80.5 & 74.3 & 77.6 & 69.7 & 62.8 & 79.6\\
				Carreira et al.~\cite{carreira2016human} 
				& 95.7 & 91.7 & 81.7 & 72.4 & 82.8 & 73.2 & 66.4 & 81.3 \\
				Newell et al.~\cite{newell2016stacked} 
				& 98.2 & 96.3 & 91.2 & 87.1 & 90.1 & 87.4 & 83.6 & 90.9\\
				Yang et al.$^*$~\cite{Yang_2017_ICCV} 
				& 98.5 & 96.7 & 92.5 & 88.7 & 91.1 & 88.6 & 86.0 & 92.0\\
				Ke et al.$^*$~\cite{Ke_2018_ECCV} 
				& 98.5 & 96.8 & 92.7 & 88.4 & 90.6 & 89.4 & 86.3 & 92.1\\
				Tang et al.$^*$~\cite{Tang_2018_ECCV} 
				& 98.4 & 96.9 & 92.6 & 88.7 & 91.8 & 89.4 & 86.2 & 92.3\\
				Xiao et al.~\cite{xiao2018simple} 
				& 98.5 & 96.6 & 91.9 & 87.6 & 91.1 & 88.1 & 84.1 & 91.5\\
				Sekii~\cite{sekii2018pose}
				& 97.9 & 95.3 & 89.1 & 83.5 & 87.9 & 82.7 & 76.2 & 88.1\\
				Zhang et al.~\cite{Zhang_2019_CVPR}
				& 98.3 & 96.4 & 91.5 & 87.4 & 90.9 & 87.1 & 83.7 & 91.1\\
				Sun et al.~\cite{Sun_2019_CVPR} 
				& 98.6 & 96.9 & 92.8 & 89.0 & 91.5 & 89.0 & 85.7 & 92.3\\
				Tang et al.$^*$~\cite{Tang_2019_CVPR} 
				& 98.7 & 97.1 & 93.1 & 89.4 & 91.9 & 90.1 & 86.7 & \textbf{92.7}\\
				Artacho et al.~\cite{artacho2020unipose} & - & - & - & - & - & - & - & \textbf{92.7}\\
				\hline
				CLNet-ResNet50 
				& 98.2 & 95.9 & 90.5 & 85.9 & 90.4 & 86.6 & 82.0 & 90.4\\
				CLNet-Hourglass
				& 98.4 & 96.6 & 92.4 & 88.4 & 90.9 & 89.4 & 84.8 & 91.9\\
				\hline
			\end{tabular}
		}
	\end{center}
	\caption{Comparisons of results on the MPII test set. "*" means using multi-scale image pyramids as input.}
	\label{table:MPII}
\end{table}

\begin{table}
	\begin{center}
		\resizebox{0.48\textwidth}{!}{
			\begin{tabular}{lcccccccc}
				\hline
				& \# Param & FLOPs & FPS & FPS' & PCKh@0.5  \\
				\hline
				
				8-stacked Hourglass, ECCV'16~\cite{newell2016stacked} 
				& 26M & 55G & 20 & 70 & 90.9\\
				PyraNet, ICCV'17$^*$~\cite{Yang_2017_ICCV} 
				& 28M & 46G & 6 & 40 & 92.0\\
				SimpleBaseline, ECCV'18~\cite{xiao2018simple} 
				& 69M & 23G & 60 & 202 & 91.5\\
				PPN, ECCV'18~\cite{sekii2018pose}
				& 16M & \underline{6G} & \textbf{388} & \textbf{728} & 88.1 \\
				HRNet, CVPR'19~\cite{Sun_2019_CVPR} 
				& 64M & 21G & 29 & 283 & \underline{92.3} \\
				FPD, CVPR'19~\cite{Zhang_2019_CVPR}
				& \textbf{3M} & 9G & 40 & 250 & 91.1 \\
				UniPose, CVPR'20~\cite{artacho2020unipose}
				& 47M & 15G & 41 & 210 & \textbf{92.7}\\
				\hline
				CLNet-ResNet50 
				& \underline{13.5M} & \textbf{5.6G} & \underline{136} & \underline{571} & 90.4\\
				\hline
			\end{tabular}
		}
	\end{center}
	\caption{Comparisons of complexity. FPS are calculated with batch size one, while FPS' are calculated using full GPU memory. "*" means using multi-scale image pyramids.}
	\label{table:speed}
\end{table}

\subsection{Experiment Setup}
\noindent
\textbf{Datasets} LSP and its extended training set provide 11k training images and 1k testing images~\cite{Johnson10}, and MPII~\cite{Andriluka_2014_CVPR} provides around 25k images with 40K person instances for single-person. MS COCO dataset~\cite{lin2014microsoft} requires localization of multi-person keypoints in the wild. COCO train2017 set includes 120K images and 150K person instances, while val2017 set and test-dev2017 set include 5K images with 6K person instances and 20K images, respectively.

\noindent
\textbf{Evaluation Protocol}
We use the Percentage of Correct Keypoints (PCK)~\cite{Andriluka_2014_CVPR} as the evaluation metric for single-person human pose estimation. The normalized distance is torso size for LSP while a fraction of the head size (referred to as PCKh) for MPII. For the MS COCO dataset, object keypoint similarity (OKS) based mAP is used as an evaluation metric.

\noindent
\textbf{Implementation Details}
The size of the input image is 256 $\times$ 256 for LSP and MPII by convention. The input size for COCO varies among experiments. The standard deviation $\sigma$ in sparse heatmaps is 16, and the threshold $\tau$ is 0.6. The loss weight of $\omega_h$ and $\omega_o$ is 0.5 and 2, respectively. Training data are augmented by shearing, scaling, rotation, flipping as reported in~\cite{newell2016stacked,Tang_2018_ECCV,xiao2018simple}. The networks are trained using PyTorch~\cite{pytorch}. We optimize the models via Adam~\cite{Adam} with a batch size of 128 for 140 epochs. The learning rate
is initialized as $1 \times 10^{-3} $ and then dropped by a factor of 10 at the 90th and 120th epochs for CLNet-ResNet. For CLNet-Hourglass, we follow the same hyper-parameters and settings in~\cite{newell2016stacked}. For top-down multi-person human pose estimation in COCO, we use the same detector as SimpleBaseline~\cite{xiao2018simple} and HRNet~\cite{Sun_2019_CVPR}. 

\begin{table}
	\begin{center}
		\resizebox{0.48\textwidth}{!}{
			\begin{tabular}{lcccccccc}
				\hline
				& Head &Sho. &Elb. &Wri. &Hip &Knee& Ank. &PCK@0.2 \\
				\hline
				Tompson et al.~\cite{tompson2014joint} 
				& 90.6 & 79.2 & 67.9 & 63.4 & 69.5 & 71.0 & 64.2 & 72.3\\
				Yang et al.$^*$~\cite{Yang_2017_ICCV} 
				& 98.3 & 94.5 & 92.2 & 88.9 & 94.4 & 95.0 & 93.7 & 93.9\\
				Tang et al.$^*$~\cite{Tang_2018_ECCV} 
				& 97.5 & 95.0 & 92.5 & 90.1 & 93.7 & 95.2 & 94.2 & 94.0\\
				Tang et al.$^*$~\cite{Tang_2019_CVPR} 
				& \textbf{98.6} & 95.4 & 93.3 & 89.8 & 94.3 & 95.7 & \textbf{94.4} & 94.5\\
				Artacho et al.~\cite{artacho2020unipose} & - & - & - & - & - & - & - & 94.5\\
				\hline
				CLNet-ResNet50 
				& \textbf{98.6} & 95.2 & 94.1 & 92.5 & 95.5 & 95.2 & 93.1 & 94.9\\
				CLNet-Hourglass 
				& 98.5 & \textbf{96.3} & \textbf{95.4} & \textbf{94.6} & \textbf{96.7} & \textbf{96.0} & 94.3 & \textbf{96.0}\\
				\hline
			\end{tabular}
		}
	\end{center}
	\caption{Comparisons of results on the LSP test set. "*" means using multi-scale image pyramids as input.}
	\label{table:LSP}
\end{table}

\begin{figure}[t]
	\begin{center}
		\includegraphics[width=1.0\linewidth]{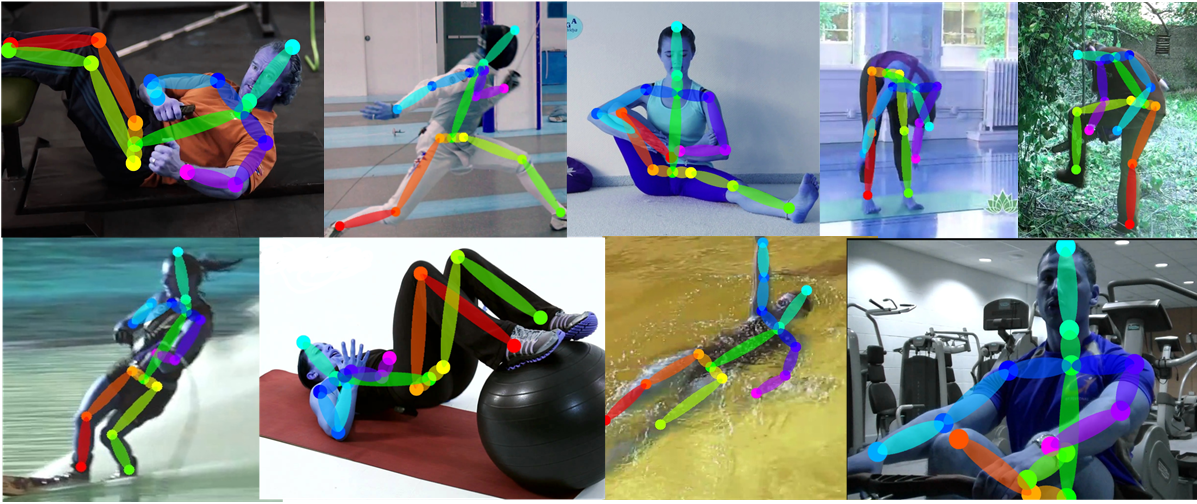}
	\end{center}
	\caption{The random qualitative results on MPII test set.}
	\label{fig:visualization}
\end{figure}

\subsection{Comparison with State-of-the-art Methods}
\label{section:cmp SOTA}
\noindent
\textbf{Results on COCO}
As shown in Table \ref{table:COCO val test}, on the COCO val2017 set, CLNet-ResNet50 achieves a 71.2 AP score with the input size 256$\times$192, and outperforms SimpleBaseline-ResNet50 by 1.14\% with near 1/2 FLOPs. CLNet-Hourglass, trained from scratch, achieves a 75.1 AP score and obtains 3.2 points improvement compared with the original Hourglass. As for the input size 384 $\times$ 288, CLNet-ResNet outperforms SimpleBaseline and CL-Hourglass outperforms the HRNet while original Hourglass is not as good as other methods. On test-dev 2017 set, CLNet-ResNet152 outperforms SimpleBaseline-ResNet152 0.4 points. CLNet-Hourglass outperforms all others with a 75.8 AP score.

\noindent
\textbf{Results on MPII}
Table \ref{table:MPII} shows our results on the MPII test. Furthermore, we compare the complexity of CLNet-ResNet50 and the most popular methods on the MPII test set in Table \ref{table:speed}. We measure the speed and latency by float-point operations (FLOPs) and frames-per-second (FPS). As shown in Table \ref{table:speed}, FPD~\cite{Zhang_2019_CVPR} has much fewer FLOPs than SimpleBaseline~\cite{xiao2018simple} but is slower, and has comparable FLOPs with PPN~\cite{sekii2018pose} but is much slower. Our method achieves an excellent trade-off between efficiency and effectiveness. Our CLNet-ResNet50 is approximately two times slower than PPN but has higher accuracy, while it has comparable accuracy but surpasses the others in terms of FPS by a large margin. Figure~\ref{fig:visualization} shows the visualization of CLnet-Resnet50 on the MPII test set.

\noindent
\textbf{Results on LSP}
Table \ref{table:LSP} shows the results of CLNets and the most popular methods on the LSP test set. Our results are the new state-of-the-art.

\begin{figure}[t]
	\begin{center}
		\includegraphics[width=0.85\linewidth]{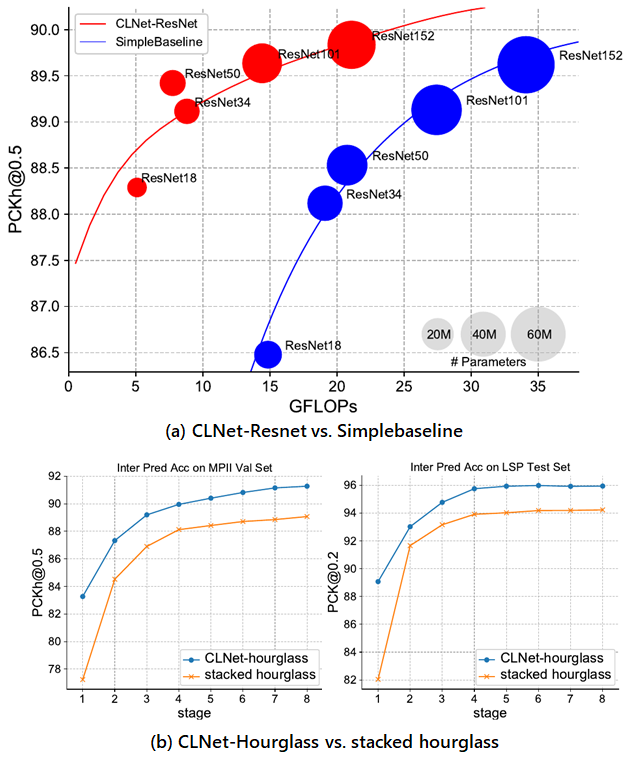}
	\end{center}
	\caption{Comparisons of accuracy at different stages of the network on our CLNet-hourglass and stacked hourglass network on MPII validation set (left) and LSP test set (right).}
	\label{fig:contrast}
\end{figure}

\subsection{Ablation Study}

\noindent
\textbf{Influence of the downsampling stride}
Different downsampling stride $S$ will make the range of short-distance regression different and the model's complexity different, which may significantly influence the result. We do experiments with different values of $S$. As Table \ref{table:heatmap size} showed, the 8 $\times$8 heatmap has the most parameters because of the vast channel numbers in the last stage, while 16$\times$16 heatmap achieves the best results with the fewest parameters.

\begin{table}
	\begin{center}
		\resizebox{0.35\textwidth}{!}{
			\begin{tabular}{c|c|c|c}
				\hline
			 	stride $S$ & Heatmap Size & Params & PCK@0.2 \\
				\hline
				4 & 64x64 & 16.8M & 92.2\\
				8 & 32x32 & 15.7M & 93.9\\
				16 & 16x16 & \textbf{13.4M} & \textbf{94.9}\\
				32 & 8x8 & 28.9M & 93.4\\
				\hline
			\end{tabular}
		}
	\end{center}
	\caption{Comparisons of results on the LSP test set with different downsampling stride $S$.}
	\label{table:heatmap size}
\end{table}

\noindent
\textbf{Influence of Loss Function}
We consider the effect of two other loss functions. The first calculates the MSE loss on offsetmaps where the corresponding position in the heatmap is the peak and keeps the heatmap loss the same as traditional Gaussian heatmap loss with a 93.2 score. The second is the loss function used in G-RMI~\cite{Papandreou_2017_CVPR}, which solves a binary classification problem for heatmap and calculates each positive location has offset the loss with a 93.7 score. Our loss function helps the models achieve the best score of 94.9.

\noindent
\textbf{Cost-effectiveness Analysis}
As shown in Figure\ref{fig:contrast} (a), we test different backbone networks in design. The performance continued to improve as the network deepens. SimpleBaseline is a baseline for effectiveness and efficiency verification of CLNet-ResNet due to the most similar architecture. CLNet-ResNet outperforms the SimpleBaseline in all backbones. Using the same backbone, CLNet-ResNet has fewer GFLOPs and parameters than SimpleBaselines because our networks do not have any up-sampling layer and the last stage with vast channels in ResNet. The gap goes from 0.2\% to 1.8\% when the backbone changes from ResNet152 to ResNet18. It demonstrates that our method can use the features more efficiently.

\noindent
\textbf{Generalization}
Our method can also be used in other popular models. We do our generalization experiments on the stacked hourglass network, named CLNet-hourglass. Although it is not very elegant compared with CLNet-ResNet, it dramatically improves performance (about 2\%) on two commonly used single-person pose estimation benchmarks, as shown in Figure \ref{fig:contrast} (b). Especially, CLNet-hourglass surpasses the original stacked hourglass network on the first stage by a large margin (relative 8.5\% on MPII validation set and 8.0\% on LSP test set), indicating our method can using features more effectively again.

\section{Conclusion}
In this paper, we has proposed the composite localization for human pose estimation, dividing the complex learning objective into two simpler ones and a combined heatmap-based and regression-based method to solve them. Besides, we have constructed two types of CLNets with different backbones and design appropriate loss functions. With fewer parameters than their plain counterparts, CLNets have achieved better average precision on three standard benchmark datasets, proving our framework's effectiveness and generalizability. We expect to optimize the CL framework and design a more elegant and powerful network in future work.

\bibliographystyle{IEEEbib}
\bibliography{icme2021template}

\end{document}